\documentclass{article}

\usepackage[nonatbib, preprint]{neurips_2024}

\usepackage[utf8]{inputenc} %
\usepackage[T1]{fontenc}    %
\usepackage{hyperref}       %
\usepackage{url}            %
\usepackage{booktabs}       %
\usepackage{amsfonts}       %
\usepackage{nicefrac}       %
\usepackage{microtype}      %
\usepackage{xcolor}         %

\usepackage{bm}
\usepackage{color}
\usepackage{xcolor}
\usepackage{multirow}
\usepackage{ifthen}
\usepackage{dsfont}
\usepackage{pifont}%
\usepackage{amsmath}
\usepackage{amssymb}
\usepackage{booktabs}
\usepackage{algorithm}
\usepackage{algorithmic}
\usepackage{graphicx}

\usepackage{nicefrac}
\usepackage{bigstrut}

\newcommand{\method}[1]{\ifthenelse{\equal{#1}{full}}{Text-Enhanced Alignment}{TEA}}

\newcommand{\vparagraph}[1]{\noindent\textbf{#1}\quad}

\usepackage{pifont}%

\def\onedot{.}
\def\eg{\emph{e.g}\onedot} 
\def\ie{\emph{i.e}\onedot} 
 
 \def\vs{\emph{vs}\onedot}

\DeclareMathOperator{\agg}{\text{Maxpool}}
\DeclareMathOperator{\mpl}{\text{Meanpool}}
\DeclareMathOperator{\fc}{\text{FC}}
\DeclareMathOperator{\softmax}{\text{Softmax}}

\newcommand{\wt}[1]{{\color{black}#1}}

\title{
 {
MLLM as Video Narrator: Mitigating Modality Imbalance in Video Moment Retrieval}
}

\author{%
    Weitong Cai
    \\
    Queen Mary University of London\\
    \texttt{weitong.cai@qmul.ac.uk} \\
  \And
  Jiabo Huang \\
  Queen Mary University of London \\
  \texttt{jiabo.huang@qmul.ac.uk} \\
  \AND
  Shaogang Gong \\
  Queen Mary University of London \\
  \texttt{s.gong@qmul.ac.uk} \\
  \And
  Hailin Jin \\
  Adobe Research\\
  \texttt{hljin@adobe.com} \\
  \And
  Yang Liu\thanks{Corresponding author} 
  \\
  WICT, Peking University \\
  \texttt{yangliu@pku.edu.cn} \\
}

\begin{document}

\maketitle

\begin{abstract}
 {
Video Moment Retrieval (VMR) aims to localize 
a specific temporal segment
within an untrimmed long video given a natural language query. 
Existing
methods often suffer from 
inadequate training annotations,
\ie,
the
sentence typically matches with a fraction of the prominent video content in
the foreground with limited wording diversity.
This intrinsic modality imbalance leaves a considerable portion of
visual information remaining unaligned with text.
It confines the cross-modal alignment knowledge within the scope of a limited text corpus,
thereby leading to sub-optimal visual-textual
modeling and poor generalizability. 
By leveraging the visual-textual understanding capability 
of multi-modal large language models (MLLM), 
in this work, 
we take an MLLM as a video narrator to generate 
plausible 
textual descriptions of the video,
thereby mitigating the modality imbalance and boosting the temporal localization.
To effectively maintain temporal sensibility for localization,
we design to get text narratives for each certain video timestamp 
and construct a structured text paragraph with time information, which is temporally aligned with the visual content.
Then we perform cross-modal feature merging between the temporal-aware narratives 
and corresponding video temporal features to produce semantic-enhanced video representation sequences for query localization. 
Subsequently, we introduce a uni-modal 
narrative-query
matching mechanism, 
which encourages the model to extract 
complementary information from contextual cohesive descriptions
for improved retrieval. 
Extensive experiments on two
benchmarks
show
the 
effectiveness 
and generalizability of our proposed method.
}
\end{abstract}

\section{Introduction}
\label{sec:intro}

Video Moment Retrieval (VMR) aims to identify moments of interest within untrimmed videos by predicting their temporal boundaries based on natural language query sentences describing specific activities~\cite{huang2021CRM, cai2022eva, gao2017tall}. This task poses a significant challenge in video understanding, requiring accurate comprehension of both visual and textual modalities and their precise alignment.

\begin{figure}[t]
\centering
\includegraphics[width=0.85
   \textwidth]{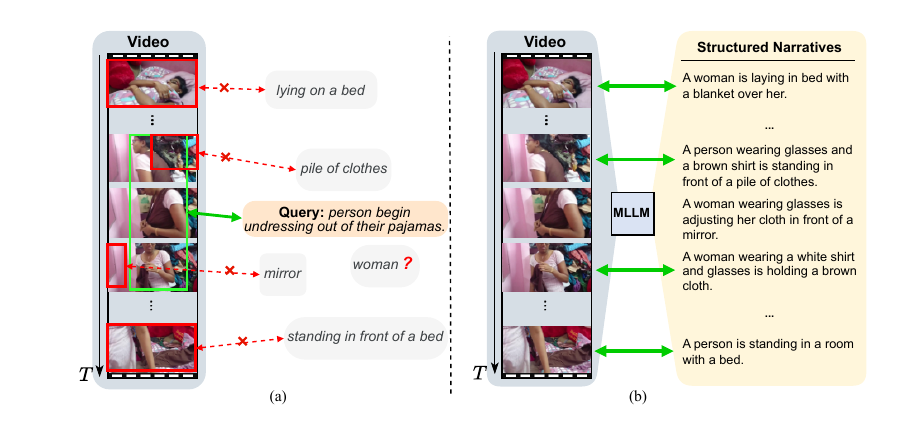}
\caption{
 {
An illustration of the intrinsic modality imbalance problem in video-query samples.
(a) Query in existing datasets solely captures a fraction of the prominent video content \textit{(semantic completeness)}
in the foreground with the limited wording \textit{diversity}, 
leaving a significant amount of visual
information unaligned with text.
(b) We leverage an MLLM as a video narrator to generate structured narratives temporally aligned with the corresponding video,
to enhance the cross-modal understanding with the rich text corpus to facilitate more accurate and generalized predictions.
}
}

\label{fig:teaser} 
\end{figure}

 {
However, 
the cross-modal alignment knowledge provided by video-query training samples 
in existing datasets~\cite{krishna2017anetcaptions,sigurdsson2016charades} 
is not always adequate
in both \textit{semantics completeness} and \textit{diversity}
to facilitate precise and generalizable moment-text correlation learning.
Specifically, firstly, in terms of \textit{semantic completeness},  
as shown in Fig.~\ref{fig:teaser}(a), 
queries describing the user's moments of interest often only capture a fraction of the video segment content, 
focusing primarily on 
partial
foreground elements rather than encompassing all the information within that moment (e.g., failing to mention details such as `mirror' and `pile of clothes'). 
Additionally, despite the query being partially associated with its corresponding moments, the extensive visual content outside the moment of interest within the same video lacks textual descriptions, 
thus unable to contribute to bridging cross-modal understanding (e.g., actions like `lying on a bed' and `standing in front of a bed'). 
 {
Secondly, in terms of \textit{diversity}, queries within a particular domain or dataset often tend to employ fixed words and phrases to convey similar or identical semantics~\cite{li2022visa,otani2020hidden}. 
}
This results in a fragile visual-textual understanding when confronted with various expressions. 
In summary, when contrasting with the richness of the video modality, the absence of a correlated text corpus, both in terms of semantic completeness and diversity, 
results in sub-optimal cross-modal learning for the VMR model. 
This intrinsic modality imbalance problem 
restricts the available multi-modal alignment information to a limited text corpus,
posing a risk of diminished generalization ability across various video-text correlations and distributions.
}

 {
Existing attempts aiming at enhancing existing textual queries to augment correlated cross-modal information are broadly categorized into two strategies.}
One entails adjusting the syntax and/or wording of the ground-truth query to generate additional pairs and seeks to improve visual-textual alignment through contrastive learning~\cite{li2023primitives,yang2023deco,hakim2023leveraging}. 
However, 
it relies on strong assumptions about the known and consistent characteristics of syntax or wording
and tailors rules to such specific traits, 
which may falter when confronted with different distributions. 
Another strategy involves leveraging manually aligned context from temporally neighboring ground-truth sentences within the same video to provide supplementation information~\cite{huang2021CRM,liu2023mesm,ramakrishnan2023naq}. 
However, depending on neighboring labeled sentences is frequently impractical in real-world scenarios with limited cross-modal annotations, thus constraining its scalability.
It is worth noting that both strategies still rely on the limited associated visual-textual information within datasets through different permutations or combinations, which perpetuates the modality imbalance problem inherent in the original annotations.

Recent advancements in Multi-modal Large Language Models (MLLM)~\cite{brown2020gpt3,chiang2023vicuna,touvron2023llama,liu2023llava,lin2023videollava} have showcased the efficacy of text prompts in facilitating textual and visual comprehension and reasoning. 
 {
Nevertheless, 
currently, many MLLMs are good at capturing the global visual semantics while hard to
associate moments with accurate timestamps directly, 
due to the compression of visual inputs and limited grounding annotations~\cite{ren2023timechat,wang2023lifelongmemory}.
}
One potential approach to leverage cross-modal knowledge is to directly generate corresponding queries from videos to create new training pairs. However, this task is non-trivial due to the ambiguity of moment boundaries~\cite{otani2020hidden}. Determining the endpoints of activities without human supervision introduces significant uncertainty and amplifies additional noise.

 {
In this work, we propose a novel MLLM-driven VMR framework
\textit{\method{full}} (\method{abbr}) to address
the modality imbalance problem 
by enhancing the correlated visual-textual knowledge.  
Instead of struggling with the difficulty of establishing correlation matching between infinite granularity visual semantics and partially relevant text description in existing dataset sample pairs,
we take an off-the-shelf MLLM as a video narrator to generate plausible textual descriptions of the video (Fig.~\ref{fig:teaser}(b)).
This choice of MLLM is driven by its capacity to enhance both the semantic completeness and diversity of the narratives
 {with prompt instructions}. 
To effectively maintain time sensibility in the generated descriptions to facilitate temporal localization,
we design to generate narratives of the video at different timestamps, 
forming a structured text paragraph
temporally aligned with the video.}
The structured paragraph converts the intricate and noisy video sequence data into cohesive log semantic summaries with time information, containing a comprehensive text corpus relevant to visual content, which is primed to narrow the cross-modal heterogeneous gap and aid in temporal moment localization.
Then we employ a 
 {video-narrative knowledge enhancement}
module to merge augmented narratives with the video feature, resulting in 
adaptively
enriched semantic-aware video representations
for query localization
by the multi-modal attention mechanism.
Subsequently, 
considering the complementary semantic description provided in the structured paragraph,
we also introduce a 
 {paragraph-query parallel interaction}
module to facilitate 
 {uni-modal}
video-query alignment. The semantic-enriched narratives play a crucial role in reducing the cross-modality heterogeneous gap, thereby improving the generalization and robustness of 
handling diverse video-text distributions.

We make three \textbf{contributions} in this work:
(1) We formulate a novel paradigm called \method{full} (\method{abbr}) to 
mitigate the modality imbalance problem and enhance generalizable moment-text associations learning in VMR.
 {
(2) We leverage an MLLM as a video narrator to construct structured textual narratives for videos. 
These narratives 
temporally aligned with videos,
serving as enriched semantic bridges to aid in cross-modal video-query alignment.}
(3) \method{abbr} provides state-of-the-art performances on various evaluations from two popular VMR benchmarks, demonstrating the effectiveness of the proposed method for enhancing generalizable visual-textual learning.

\section{Related Works}
\label{sec:relate}

\noindent\textbf{Video Moment Retrieval (VMR).}
 {
VMR~\cite{gao2017tall,anne2017MCN}, also known as natural language video localization
or video grounding, 
requires fine-grained temporal sensibility to associate moments and queries.
}
Proposal-based 
methods~\cite{anne2017MCN,gao2017tall,ge2019mac,zhang20202dtan,zhang2019exploiting,wang2021mmn}
generate candidate video segments, aggregate all the frames with a video segment
and align them holistically with the query.
Another 
paradigm 
is proposal-free boundary identification,
aiming to directly regress the temporal coordinates 
of the target moments~\cite{yuan2019find,lu2019debug,chen2020learning,chen2020hierarchical,chen2021drft}
or predict the per-frame probabilities of being the start and 
end points~\cite{hao2022query,zhang2020vslnet,zhang2021natural,nan2021ivg,lei2020tvr}.
And some works tried to retrieve moments 
by the proposal-based and proposal-free strategies jointly~\cite{wang2020dual,wang2021smin,xiao2021boundary,huang2022emb}.
 {
All of them focused on learning fine-grained visual-textual correlation from the datasets
while suffering from the modality imbalance problem.
}

\vparagraph{Modality Imbalance in VMR.} 
 {
In existing VMR datasets, 
queries describing the user's moments of interest often only capture a fraction of the video content spatially and temporally. 
This inherent modality imbalance results in suboptimal learning of the association between moments and text.
To enhance the query and promote correlated cross-modal learning, 
several methods~\cite{yang2023deco, li2023primitives, hakim2023leveraging}
customized the rules to very specific wording and syntax characteristics,
 {
\eg, DeCo~\cite{yang2023deco} decomposed and re-combined query elements in multiple granularities, 
}
potentially reducing effectiveness when confronted with unseen distributions diverging from the tailored trained data.
\cite{ding2021support} constructed a support set, considering the simultaneous presence of certain visual entities 
but still ignoring unrelated semantics in the vision.
 {
Some works~\cite{liu2023mesm, ramakrishnan2023naq} complemented the query semantics from contexted sentences, 
\eg, MESM~\cite{liu2023mesm} discussed the word and segment-level imbalances and added prior knowledge from neighbor queries in the same video.
However, they are not always realistic in real-world scenarios where the video-text annotation is limited.
}
In contrast to the existing approaches, which rely on limited visual-textual information within datasets through various permutations or combinations, we utilize an MLLM to generate video narratives, thereby enhancing both their semantic richness and diversity, 
to aid in cross-modal alignment between video and query.
}

\vparagraph{Large Language Models in Video Understanding.}
Recent developments in large language models~\cite{brown2020gpt3,chiang2023vicuna,touvron2023llama,chiang2023vicuna} and large-scale vision-language pretraining~\cite{radford2021CLIP,li2023blip2, wang2022internvideo}
have underscored the abilities of MLLMs~\cite{liu2023llava,achiam2023gpt4,hu2024gensam} to understand visual and textual information. 
 {
Upon these, several works~\cite{li2023videochat,lin2023videollava,zhang2023videollama} trained MLLMs for video inputs but still failed to provide meaningful temporal localization predictions~\cite{wang2023lifelongmemory}.
To pursue better fine-grained video understanding,
\cite{ren2023timechat} used a Q-former to fine-tune
a time-sensitive MLLM with explicit time information for video reasoning. 
Further, \cite{wu2023cap4video} used GPT to
generate captions for trimmed videos to facilitate text-video
retrieval from a set of videos. 
\cite{wang2023lifelongmemory} transformed videos into captions and asked GPT to predict moment boundaries directly. 
However, due to the information loss in video input pre-processing and limited grounding annotations, all these approaches were still struggling to get accurate moment endpoints for localization.
Furthermore, some attempts~\cite{luo2023vdi,lei2021momentdetr} embedded vision-language pertaining knowledge in feature space for localization but still struggled with cross-modal alignment precision in heterogeneous domains.
In this work, we introduce the multi-modal understanding capacity of MLLM to a VMR pipeline 
by generating the text narratives corresponding to videos.
Different from~\cite{wu2023cap4video},  videos in VMR are untrimmed and often unscripted~\cite{huang2022emb}, 
which brings more challenges in understanding fine-grained video-text alignment knowledge and conducting temporal segment semantic searching.
We carefully form a structured paragraph temporally aligned with videos at certain timestamps, bringing rich text corpora correlated with videos for cross-modal understanding.
}

\section{Methods}
\label{sec:method}

\begin{figure*}[t]
\centering
\includegraphics[width=1.0
   \textwidth]{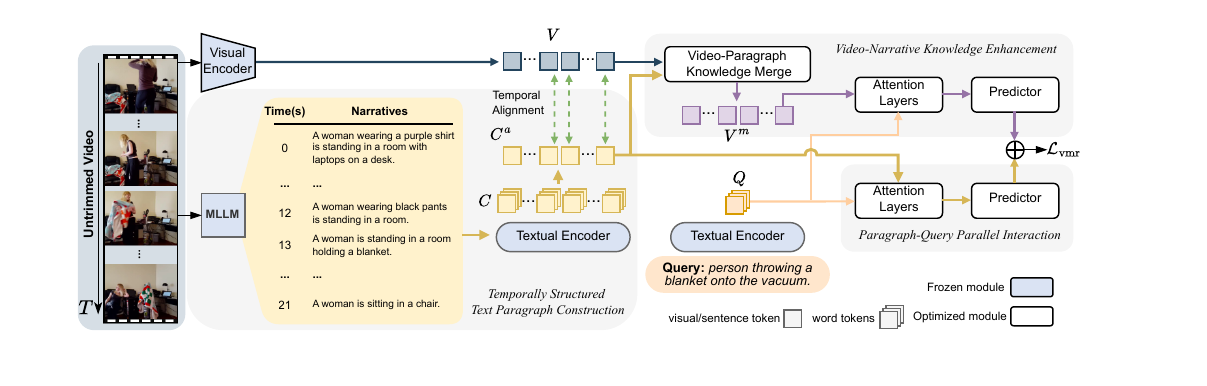}
\vspace{-1mm}
\caption{An overview of our \textit{\protect\method{full}} 
(\protect\method{abbr})
model.
\wt{
We take an offline MLLM as a video narrator to generate a structured narrative paragraph $C^a$ that is temporally aligned with the input video snippet feature sequences $V$. 
\protect\method{abbr} performs 
video-narrative
knowledge enhancement 
to acquire more discriminative text-enhanced video representations.
Parallelly, we conduct a paragraph-query interaction module to complement context understanding and promote more generalizable predictions.
}
}
\label{fig:framework}
\end{figure*}

\subsection{Problem Definition}
Given 
an untrimmed video $V$ with $T$ duration, 
and a natural language query sentence 
$Q$ 
that reflects the user's interests in specific temporal and visual parts,
the object of video moment retrieval is to semantically align the query to the target video moment segment by predicting its start and end timestamps $(\tau^s, \tau^e)$. 
It is challenging to get a semantic understanding of both visual and textual inputs and then align them to localize accurately the temporal boundaries of a certain motion behavior.

Considering the query describes only part of the video temporal information and part of the visual semantics in the target moment segment, 
it is hard for the model to understand
visual-textual correlation knowledge comprehensively
under the lack of text corpus in both semantics and syntax.
In this work, we study the 
 {modality imbalance problem}
by proposing a \textit{\method{full}} (\method{abbr}) model (Fig.~\ref{fig:framework}).
\method{abbr} first generates a structured text paragraph that is temporally aligned with the input video $V$ from an offline multi-modal large language model. The choice of MLLM is driven by its capacity to enhance both the semantic completeness and diversity of the narratives. Subsequently, \method{abbr} employs the video-narrative knowledge enhancement module to acquire more discriminative video representations enriched with augmented narrative semantics. 
Expanding upon this, we introduce 
a  
 {paragraph-query parallel interaction module}
to facilitate cross-modal video-query alignment, addressing the imbalance problem and cross-modal heterogeneous gap. This design aims to improve the generalization and robustness of the VMR model in managing diverse video-text distributions.
Adopting the convention~\cite{nan2021ivg,wang2021smin, luo2023vdi},
we represent video snippet features with a pretrained CNN as 
$V = \{s_i\}^{L^v}_{i=1}$ composed of $L^v$ snippets where each captures a non-overlap time period $[t_{s_i}^s, t_{s_i}^e]$, 
and the query sentence by the GloVe embeddings as 
$Q = \{w_i\}^{L^q}_{i=1}$ 
with $L^q$ words.

\subsection{Temporally Structured Text Paragraph Construction}
 {
Recent developments in large language models~\cite{achiam2023gpt4,liu2023llava} and large-scale vision-language pretraining~\cite{radford2021CLIP,li2023blip2}
have underscored the extraordinary abilities of MLLMs to understand both visual and textual information. 
Upon this, to mitigate the modality imbalance in video-query samples, 
we
use one off-the-shelf MLLM as a video narrator to 
generate multifaceted and diverse caption descriptions 
related to the video content.
To effectively maintain temporal sensibility in the generated descriptions for moment temporal localization,
in this section, 
we utilize one offline MLLM to convert one video to multiple text descriptions at different timestamps and then construct a structured paragraph.}

Given the input video $V$, we utilize a pre-trained $\mathrm{MLLM}(\cdot)$ to transcribe the raw visual data into a list of narrative descriptions. 
Specifically,
$V$ is firstly sampled to image frames $\{f_i\}^{L^f}_{i=1}$ at fixed
time intervals $m$, where $L^f$ is the number of frames. Then we
instruct $\mathrm{MLLM}(\cdot)$ with prompt $P =$ 
``This is one image frame sampled from a video. Please caption this frame in two or three sentences, to describe this frame with some details but without any analysis.'' 
to yield the text narrative of each frame $f_i$ at a certain timepoint $t_{f_i}$, as follows:
\begin{equation}
c_i = \mathrm{MLLM}(f_i, P),
\label{eq:caption}
\end{equation}
where $c_i$ is the response answer as the text description of $f_i$, 
capturing the video's narrative 
that correlates the visual semantics in a more accessible and explainable format.
 {
Then all the text narratives are concatenated in chronological order 
to construct a text paragraph $C$ with time information as:}
\begin{equation}
C = \{t_{f_i}\!\!: \: \, c_i\}_{i=1}^{L^f}.
\end{equation}
In practice,  the sampling rates of individual videos may vary~\cite{zhang20202dtan, wang2021mmn,zhang2020vslnet,hao2022shuf} 
due to the diversity of video acquisition and codec processing,
leading to potential inconsistent sampling rates between videos and the constructed corresponding paragraphs.
In this case, for fine-grained visual-textual semantic matching, 
we design to align the paragraph with the video features sequence temporally.
Similar to the merge operation on video features in~\cite{zhang2020vslnet}, 
we mean-pooling the  {neighbor} narrative features whose time points fall into the same snippet period as follows:
\begin{equation}
C^a = \{c^{a}_k\}_{k=1}^{L^v} = \mpl(\{\text{Sent}(\mathcal{F}_T(c_m)), \text{where} \: t_{f_m} \in [t_{s_k}^s, t_{s_k}^e]\}),
\label{eq:para_align}
\end{equation}
where $\mathcal{F}_T$ is the frozen text feature extractor to get word-level embeddings
and $\text{Sent}(\cdot)$ is to generate sentence-level features by averaging.
After the alignment, 
the structured text paragraph feature $C^a$ are semantically matched with the input video $V$ on the time dimension.
This matching process facilitates the conversion of intricate and noisy video visual sequence data into a cohesive log of semantic summaries
 {with time sensibility,} 
containing a valuable and comprehensive text corpus relevant to the visual content. This corpus is primed to aid in temporal moment localization.
 {
Moreover, 
the design of merging multiple neighbor narratives
from coherent similar visual content
to align with the video granularity, 
will also improve the robustness to resist potential noises from MLLM's output.}
For simplicity, we reuse $C$ to represent the temporally aligned structured paragraph features $C^a$.

\subsection{Video-Narrative Knowledge Enhancement}
The structured paragraph corresponding to the video contains the text narrative summary 
in different time periods. 
Due to the intrinsic characteristic
of sharing the same semantic space with the query $Q$, 
the paragraph $C$ bridges the semantic understanding between the abundant temporal visual data with abstract textual query information.
In this section, we leverage the complementarity between videos and augmented text narratives to cultivate more discriminative text-enhanced video sequence representations. These representations help to narrow the cross-modal heterogeneous gap, thereby aligning video moments with the query $Q$.

\vparagraph{Video-paragpagh knowledge merging.}
To model weighted combinations of the video snippet and the text narratives, 
we concatenate each snippet feature $s_i$ with the corresponding paragraph text embedding $c_i$ 
on the hidden feature dimension and utilize a learnable multi-layer perceptron (MLP) to get the information-merged video snippet feature $s^m_i$ as follows:
\begin{equation}
s^m_i = \text{MLP}(s_i \mathbin\Vert  c_i),
\label{eq:merge}
\end{equation}
where $(\cdot \mathbin\Vert \cdot)$ denotes the hidden-feature-wise concatenation.
With the merge operation, the video feature $s^e_i$ reduces the redundant noise in the visual input and complements the augmented text narrative information which is naturally more compatible with the query.
Then we get the information-merged video sequence $V^m$:
\begin{equation}
V^m = \{s^m_i\}_{i=1}^{L^v}.
\end{equation}

\vparagraph{Query-attended knowledge enhancement.}
After getting the information-combined video features from the structured paragraph narratives, 
we facilitate knowledge enhancement by deploying attentive encoding~\cite{vaswani2017transformer,cai2022eva} for both visual and textual representations to analyze the correlation among elements in both.
 {
In one attention unit $\mathcal{A}$ on sequence analysis,
given a target sequence $X^t \in \mathbb{R}^{L^t \times d}$ with $L^t$ elements
and a reference $X^r \in \mathbb{R}^{L^r \times d}$ with $L^r$ length,
}
$\mathcal{A}(X^t, X^r)$ 
attends $X^t$ using $X^r$ as follows: 
\begin{equation}
\small
\begin{split}
\mathcal{R}(X^t, X^r)&= \softmax(\fc(X^t)\fc(X^r)^\top/\sqrt{d})
\in \mathbb{R}^{L^t\times L^r}, 
\\
\mathcal{A}(X^t, X^r) &= \fc(X^t + \mathcal{R}(X^t, X^r)\fc(X^r))
\in \mathbb{R}^{L^t\times d}.
\label{eq:attention_unit}
\end{split}
\end{equation}
The attention unit $\mathcal{A}(X^t, X^r)$ 
in Eq.~\eqref{eq:attention_unit}
is parameterized by four independent fully connected (FC) layers.
And we also conduct a guided mechanism~\cite{huang2022emb}
on video features:
\begin{equation}
\small
\label{eq:guided-attention}
\hat V^m = \text{Conv2D}(\{V^m, \{\agg(\{s^m_i\}_{i=1}^t)\}_{t=1}^{L^v},
\{\agg(\{s^m_i\}_{i=t}^{L^v})\}_{t=1}^{L^v} \}). 
\end{equation}
Then we promote self- and cross-attention 
for context exploration 
and knowledge enhancement
by:
\begin{equation}
    V^e, Q^e = \text{Attn}(\hat V^m, Q),
    \label{eq:attention}
\end{equation}
where in the $\text{Attn}(X, Y)$ function:
\begin{equation}
X \leftarrow \mathcal{A}(X, X), X \leftarrow \mathcal{A}(X, Y); \,\,\,\,\,
Y \leftarrow \mathcal{A}(Y, Y), Y \leftarrow \mathcal{A}(Y, X).
\label{eq:attention_step}
\end{equation}
 {After both merging and multi-modal attention operations,
we combine generated narratives with video features and adaptively enrich semantic-aware video presentations for query-relevant localization.}

\vparagraph{Endpoint prediction.}
 {
With the knowledge-enhanced video and query features$(V^e, Q^e)$,
our \method{abbr} model is ready to benefit existing VMR predictors.
Here, we take 
the state-of-the-art span-based predictor~\cite{huang2022emb} 
as an example
to get per-snippet scores of being the start and end time points $(p^s_v, p^e_v)$:
}
\begin{equation}
(p^s_v, p^e_v) = \text{Predictor}(V, Q) = 
  \softmax(\text{LSTM}(\bar{V}\odot h)),
\label{eq:predict}
\end{equation}
where
\begin{equation}
\small
\begin{split}
&h = \sigma(\text{Conv1D}(\bar{V} \Vert q)); 
\\
&\bar{V} = H(V, Q) = \fc(V\Vert X^{v2q}\Vert V\odot X^{v2q}\Vert V \odot X^{q2v});
\,\, 
\textrm{and} \\ 
&R = \fc(V)\fc(Q)^\top/\sqrt{d},\ 
X^{v2q} = {R^r}Q,\ 
X^{q2v} = R^r{R^c}^\top V;
\end{split}
\label{eq:predict_detail}
\end{equation}
$q$ is the sentence-level query feature by the weighted sum of words~\cite{bahdanau2014neural},
$\sigma$ is the sigmoid function,
$\odot$ denotes Hadamard Product 
and $(\cdot \Vert \cdot)$ is concatenation.
$R^r$, $R^c$ are deployed row and column-wise softmax operation on $R$.

\subsection{Paragraph-Query Parallel Interaction}
The generated structured paragraph $C$ is a cohesive log of semantics summaries, containing a comprehensive text corpus relevant to the visual content. 
It also may highlight some content as a complementation for query localization.
For example, if the query would like to find `another person', the different narratives of the two people in the paragraph
(`a man with pink shirt' \vs `a woman with blur dress') 
will provide a clear guide for the `another' in the uni-modal understanding.
Considering that the structured paragraph $C$ shares the same semantic space with the query,
and the paragraph also has temporal discrimination to guide the moment localization, 
in this section, we conduct a paragraph-query interaction parallelly 
to enhance the semantic alignment in the text-text interaction space as a complement.
Specifically, given the structured paragraph $C$ and the query $Q$, we first apply the similar attention interaction in Eq.~\eqref{eq:attention} independently to get the attention attended paragraph and query features as follows:
\begin{equation}
C, Q \leftarrow \text{Attn}(C, Q).
\label{eq:pqattn}
\end{equation}
And then the start and end point scores are calculated as:
\begin{equation}
(p^s_c, p^e_c) = \text{Predictor}(C, Q).
\label{eq:pqpredict}
\end{equation}
Then the final start and end point scores $(p^s, p^e)$ are enhanced by weighting both the video-query prediction and paragraph-query prediction:
\begin{equation}
p^s = p^s_v + \alpha p^s_c, \;
p^e = p^e_v + \alpha p^e_c,
\label{eq:enhanced_predict}
\end{equation}
where $\alpha$ is the weighted hyper-parameter.

\subsection{Model Training}
\label{subsec:train}
In the training stage, we follow~\cite{huang2022emb}
to expand label boundaries $(\tau^s, \tau^e)$ 
into candidate endpoint sets $(\tilde{\bm{\tau}}^s, \tilde{\bm{\tau}}^e)$ 
by an auxiliary proposal ranking.
Then the VMR model retrieval loss is computed as:
\begin{equation}
\mathcal{L}_{vmr} = -\text{log}(\sum_{i \in \tilde{\bm{\tau}}^s}p_i^s)
-\text{log}(\sum_{i \in \tilde{\bm{\tau}}^e}p_i^e),
\end{equation}
and
we highlight foreground video content by learning $h$ in Eq.~(\ref{eq:predict_detail}) as:
\begin{equation}
\mathcal{L}_h = \text{BCE}(y^h, h), \,\, y_i^h = \mathds{1}[\min(\tilde{\bm{\tau}}^s)  \leq i \leq \max(\tilde{\bm{\tau}}^e)].
\end{equation}
 {
Then the overall loss of \method{abbr} is then formulated as:
\begin{equation}
\footnotesize
\mathcal{L} = 
\mathcal{L}_\text{vmr} +
\lambda \mathcal{L}_\text{h},
\label{eq:loss} 
\end{equation}
where $\lambda$ is a loss hyper-parameter. }

\section{Experiments}
\label{sec:experiment}

\vparagraph{Datasets.}
Experiments were conducted on two popular VMR benchmark datasets:
\textbf{(1)}
Charades-STA~\cite{gao2017tall} 
is built upon the Charades dataset~\cite{sigurdsson2016charades}, 
which is mainly about indoor activities,
for video captioning and action recognition.
The work of \cite{gao2017tall} adapted the dataset to the VMR task
by collecting the query annotations. 
\textbf{(2)}
ActivityNet-Captions~\cite{krishna2017anetcaptions}
is built on~\cite{caba2015activitynet} for the dense video captioning task, 
which is a large-scale dataset of human activities based on YouTube videos.
ActivityNet-Captions is a much larger dataset compared to Charades-STA, with more sample pairs (71.9k \vs 16.1k) 
and more diverse information. 
Table~\ref{tab:dataset} illustrates their quite different data characteristics.
The average length of both moments and videos in ActivityNet-Captions 
are much longer than those in Charades-STA, 
which leads to highly varied semantics richness in vision.
Regarding the query, the descriptions in Charades-STA are much shorter,
many consisting only of simple subject-verb-object structures with limited vocabularies (1.3k \vs 12.5k), 
and are sometimes incomplete.

\begin{table}[b]
\footnotesize
\centering
\caption{Statistics of benchmark datasets.
  }
\resizebox{\linewidth}{!}{%
\begin{tabular}{lccccccp{18.5em}}
\toprule
\multirow{2}[4]{*}{Dataset} & \multirow{2}[4]{*}{\#video} & \multirow{2}[4]{*}{\#moment} & \multicolumn{2}{c}{avg. len. (sec)} & avg. len. (wrd) & \multirow{2}[4]{*}{Vocab. Size} & \multicolumn{1}{c}{\multirow{2}[4]{*}{Query Example}} \\
\cmidrule(lr){4-5} \cmidrule(lr){6-6}        &       &       & moment & video & query &       & \multicolumn{1}{c}{} \\
\midrule
\midrule
Charades & 6,672 & 16,128 & 8.1   & 30.6  & 7.2   & 1.3k  & \multicolumn{1}{l}{person they put on their shoes.} \\
\midrule
\multirow{3}[2]{*}{Anet} & \multirow{3}[2]{*}{14,926} & \multirow{3}[2]{*}{71,957} & \multirow{3}[2]{*}{36.2} & \multirow{3}[2]{*}{117.6} & \multirow{3}[2]{*}{14.8} & \multirow{3}[2]{*}{12.5k} & \multicolumn{1}{l}{As the woman moving the dolphins started} \\
      &       &       &       &       &       &       & \multicolumn{1}{l}{to swim with her, the dolphins' fins are vis-} \\
      &       &       &       &       &       &       & ible as they swim up and down. \\
\bottomrule
\end{tabular}%

}
\label{tab:dataset}
\end{table}

\vparagraph{Dataset generalizablity evaluation splits.} 
To measure the visual-textual matching performance and generalization ability,
there are several distinct splits according to different aspects of testing 
in the two benchmark datasets.
CD-Test-ood~\cite{yuan2021ood}
is proposed to effectively 
evaluate the video-query alignment under temporal annotation bias, especially the retrieval accuracy of specific key instances and verbs in the sentence, 
by introducing unseen locations in the test set.
CG-Novel-word~\cite{li2022visa} is designed for testing the generalization capability of unseen words.
And CG-Novel-composition~\cite{li2022visa} is to assess the compositional capability by different query wording and compositions.
All three are widely used dataset splits in VMR to evaluate the cross-modal understanding quality. 
Given the diverse facets targeted by these splits,
each poses distinct challenges.

\vparagraph{Performance metrics.}
Following the convents~\cite{hao2022shuf,zhang2020vslnet,cai2022eva}
to fairly measure 
results,
we use ``$\text{IoU}@m$''
to calculate the percentage of the top predicted moment having Intersection over Union (IoU) with ground truth larger than $m$,
and also adopt ``mIoU'' to represent the average IoU over all testing samples.
We report the results as $m \in \{0.5,0.7\}$
for fair comparison following~\cite{hao2022shuf,yang2023deco}.

\vparagraph{Implementation details.}
For video modality, 
we used the features provided by~\cite{huang2022emb}.
GloVe embeddings~\cite{pennington2014glove} were utilized 
as the word-level
feature embeddings.
Videos were downsampled to 128 frames at most by max-pooling
and zero-padded the shorter ones.
The dimension of all the hidden layers was fixed at 128, 
and the number of attention head in Eq.~\eqref{eq:attention_unit} was 8
followed by layer normalization and 0.2 dropout rate.
The frame interval $m$ was 1 second.
We took a pre-trained \texttt{LLaVA-v1.5-13b}~\cite{liu2023llava} as the multi-modal large language model 
to generate the text descriptions.
The \method{abbr} model 
was trained 100 epochs by the Adam optimizer 
using a linearly decaying learning rate of $0.0005$ and gradient clipping of $1.0$
with a batch size of 16.
 {
The hyper-parameter 
in Eq.~(\ref{eq:loss}) 
was empirically set as 
$\lambda =5$.
}
And the weighted hyper-parameter $\alpha$ in~\eqref{eq:enhanced_predict} was set to 0.5.
All experiments
were implemented by PyTorch,
and run on a single NVIDIA A100 40G GPU.

\begin{table}[t]

\begin{minipage}{\textwidth}
\footnotesize
\centering
\caption{Comparisons with SOTAs on Charades-STA.
$\dagger$ denotes the reproduced results 
under the strictly identical setups
using the code from authors.
Best results are in \protect\textbf{bold}.
The grey row indicates using additional ground-truth descriptions in the same video.
  }
\vspace{-1mm}
\resizebox{1.\textwidth}{!}{
\renewcommand{\arraystretch}{0.91}
\begin{tabular}{l|c|ccc|ccc|ccc}
\hline
\multirow{2}[4]{*}{Method} & \multirow{2}[4]{*}{Year} & \multicolumn{3}{c|}{CD-Test-ood} & \multicolumn{3}{c|}{CG-Novel-word} & \multicolumn{3}{c}{CG-Novel-composition} \bigstrut\\
\cline{3-11}      &       & IoU@0.5 & IoU@0.7 & mIoU  & IoU@0.5 & IoU@0.7 & mIoU  & IoU@0.5 & IoU@0.7 & mIoU \bigstrut\\
\hline
2D-TAN~\cite{zhang20202dtan} & 2020  & 35.88 & 13.91 & 34.22 & 29.36 & 13.21 & 28.47 & 30.91 & 12.23 & 29.75 \bigstrut[t]\\
LGI~\cite{mun2020lg} & 2020  & 42.90 & 19.29 & 39.43 & 26.48 & 12.47 & 27.62 & 29.42 & 12.73 & 30.09 \\
VSLNet~\cite{zhang2020vslnet} & 2020  & 34.10 & 17.87 & 36.34 & 25.60 & 10.07 & 30.21 & 24.25 & 11.54 & 31.43 \\
DRN~\cite{zeng2020dense} & 2020  & 31.11 & 15.17 & 23.05 & -     & -     & -     & -     & -     & - \\
DCM~\cite{yang2021dcm} & 2021  & 45.47 & 22.70 & 40.99 & -     & -     & -     & -     & -     & - \\
VISA~\cite{li2022visa} & 2022  & -     & -     & -     & 42.35 & 20.88 & 40.18 & 45.41 & 22.71 & 42.03 \\
Shuffling~\cite{hao2022shuf} & 2022  & 46.67 & 27.08 & 44.30 & -     & -     & -     & -     & -     & - \\
EMB$^\dagger$~\cite{huang2022emb} & 2022  & 51.97 & 31.08 & 48.51 & 48.92 & 28.92 & 45.18 & 43.61 & 24.58 & 41.07 \\
Primitives~\cite{li2023primitives} & 2023  & -     & -     & -     & 50.36 & 28.78 & 43.15 & 46.54 & 25.10 & 40.00 \\
DeCo~\cite{yang2023deco} & 2023  & -     & -     & -     & -     & -     & -     & \textbf{47.39} & 21.06 & 40.70 \\
BM-DETR~\cite{jung2023bmdetr} & 2023  & 49.32 & 27.12 & 45.18 & -     & -     & -     & -     & -     & - \\
VDI~\cite{luo2023vdi} & 2023  & -     & -     & -     & 46.47 & 28.63 & 41.60 & -     & -     & - \\
\textcolor[rgb]{ .682,  .667,  .667}{MESM~\cite{liu2023mesm}} & \textcolor[rgb]{ .682,  .667,  .667}{2024} & \textcolor[rgb]{ .682,  .667,  .667}{-} & \textcolor[rgb]{ .682,  .667,  .667}{-} & \textcolor[rgb]{ .682,  .667,  .667}{-} & \textcolor[rgb]{ .682,  .667,  .667}{50.50} & \textcolor[rgb]{ .682,  .667,  .667}{33.67} & \textcolor[rgb]{ .682,  .667,  .667}{46.20} & \textcolor[rgb]{ .682,  .667,  .667}{46.19} & \textcolor[rgb]{ .682,  .667,  .667}{26.00} & \textcolor[rgb]{ .682,  .667,  .667}{41.40} \\
\method{abbr} (Ours) & 2024  & \textbf{54.28} & \textbf{33.04} & \textbf{50.28} & \textbf{50.94} & \textbf{32.66} & \textbf{47.34} & 45.00 & \textbf{27.75} & \textbf{42.09} \bigstrut[b]\\
\hline
\end{tabular}%

}
\label{tab:sota_ood_charades}
\end{minipage}

\vspace{1.5mm}
\begin{minipage}{\textwidth}
\footnotesize
\centering
\caption{Comparisons with SOTAs on ActivityNet-Captions.
$\dagger$ denotes the reproduced results 
under the strictly identical setups
using the code from authors.
Best results are in \protect\textbf{bold}.
  }
\vspace{-1mm}
\resizebox{1.\textwidth}{!}{
\renewcommand{\arraystretch}{0.91}
\begin{tabular}{l|c|ccc|ccc|ccc}
\hline
\multirow{2}[4]{*}{Method} & \multirow{2}[4]{*}{Year} & \multicolumn{3}{c|}{CD-Test-ood} & \multicolumn{3}{c|}{CG-Novel-word} & \multicolumn{3}{c}{CG-Novel-composition} \bigstrut\\
\cline{3-11}      &       & IoU@0.5 & IoU@0.7 & mIoU  & IoU@0.5 & IoU@0.7 & mIoU  & IoU@0.5 & IoU@0.7 & mIoU \bigstrut\\
\hline
2D-TAN~\cite{zhang20202dtan} & 2020  & 22.01 & 10.34 & 28.31 & 23.86 & 10.37 & 28.88 & 22.80 & 9.95  & 28.49 \bigstrut[t]\\
LGI~\cite{mun2020lg} & 2020  & 23.85 & 10.96 & 28.46 & 23.10 & 9.03  & 26.95 & 23.21 & 9.02  & 27.86 \\
VSLNet~\cite{zhang2020vslnet} & 2020  & 20.03 & 10.29 & 28.18 & 21.68 & 9.94  & 29.58 & 20.21 & 9.18  & 29.07 \\
DCM~\cite{yang2021dcm} & 2021  & 22.32 & 11.22 & 28.08 & -     & -     & -     & -     & -     & - \\
VISA~\cite{li2022visa} & 2022  & -     & -     & -     & 30.14 & 15.90 & 35.13 & 31.51 & 16.73 & 35.85 \\
Shuffling~\cite{hao2022shuf} & 2022  & 24.57 & 13.21 & 30.45 & -     & -     & -     & -     & -     & - \\
EMB$^\dagger$~\cite{huang2022emb} & 2022  & 27.72 & 14.03 & 31.25 & 31.97 & 15.82 & 34.87 & 31.49 & 16.00 & 35.31 \\
Primitives~\cite{li2023primitives} & 2023  & -     & -     & -     & 30.15 & 14.97 & 32.14 & 30.80 & 15.39 & 33.18 \\
DeCo~\cite{yang2023deco} & 2023  & -     & -     & -     & -     & -     & -     & 28.69 & 12.98 & 32.67 \\
VDI~\cite{luo2023vdi} & 2023  & -     & -     & -     & 32.35 & 16.02 & 34.32 & -     & -     & - \\
\method{abbr} (Ours) & 2024  & \textbf{27.98} & \textbf{14.45} & \textbf{31.85} & \textbf{32.89} & \textbf{16.79} & \textbf{35.17} & \textbf{32.97} & \textbf{17.49} & \textbf{36.24} \bigstrut[b]\\
\hline
\end{tabular}%

}
\label{tab:sota_ood_anet}
\vspace{2mm}
\end{minipage}

\begin{minipage}{\textwidth}
\small
\centering
\begin{minipage}[t]{0.5\textwidth}
\scriptsize
\centering
\renewcommand{\arraystretch}{0.8}
\setlength{\tabcolsep}{12.5pt}
\caption{Video-paragraph merging choices}
\vspace{-2mm}
\begin{tabular}{l|cc|c}
\hline
\multirow{2}[4]{*}{Implementation} & \multicolumn{2}{c|}{IoU@m} & \multirow{2}[4]{*}{mIoU} \bigstrut\\
\cline{2-3}      & 0.5   & 0.7   &  \bigstrut\\
\hline
Video-only & 51.97 & 31.08 & 48.51 \bigstrut[t]\\
Add   & 52.92 & 32.09 & 49.50 \\
Attention & 52.98 & 31.82 & 48.83 \\
Concat+MLP & \textbf{54.28} & \textbf{33.04} & \textbf{50.28} \bigstrut[b]\\
\hline
\end{tabular}%

\label{tab:abla-merge}
\end{minipage}\hfill 
\begin{minipage}[t]{0.48\textwidth} 
\scriptsize
\centering
\renewcommand{\arraystretch}{1.05}
\setlength{\tabcolsep}{12pt}
\caption{Different text generator choices
  }
\vspace{-1mm}
\begin{tabular}{l|cc|c}
\hline
\multirow{2}[4]{*}{Text Generator} & \multicolumn{2}{c|}{IoU@m} & \multirow{2}[4]{*}{mIoU} \bigstrut\\
\cline{2-3}      & 0.5   & 0.7   &  \bigstrut\\
\hline
BLIP~\cite{li2023blip2} & 53.93 & 32.65 & 49.53 \bigstrut[t]\\
LLaVA~\cite{liu2023llava} & \textbf{54.28} & \textbf{33.04} & \textbf{50.28} \bigstrut[b]\\
\hline
\end{tabular}%

\label{tab:abla-captioner}
\end{minipage}
\end{minipage}

\vspace{-5mm}
\end{table}

\vspace{-1mm}
\subsection{Comparisons with the State-of-the-arts}
\vspace{-1mm}
To validate the generality and effectiveness of our proposed \method{abbr},
we compared TEA with existing methods on all three data splits
in both 
Charades-STA 
and ActivityNet-Captions
datasets.
The quantitative results are shown in Tables~\ref{tab:sota_ood_charades} and~\ref{tab:sota_ood_anet}, respectively.
One can see that TEA outperforms the SOTA methods by a significant margin 
on most metrics in all three tests of both Charades-STA 
and ActivityNet-Captions datasets. TEA can generate comparable
results compared with MESE~\cite{liu2023mesm} whilst MESE also
  utilizes additional ground-truth sentences in one video to complement context. 
The performances in different visual-textual understanding challenges
demonstrate the effectiveness of our text-enhanced alignment model. 
With temporally structured paragraphs, \method{abbr} can bridge the gap between video and text modalities and address the modality imbalance problem to get a more accurate and generalizable cross-modal understanding. 
Given the larger amount of samples, richer syntax/wording expressions, and longer video/moment durations, 
ActivityNet-Captions poses additional challenges in video-query semantic understanding. 
\method{abbr} still achieves promising performances in all evaluations in Table~\ref{tab:sota_ood_anet}.
Although 
some methods
get higher $\text{IoU}@0.5$ results on a certain split in one dataset by tailor-made design for specific tests, 
\method{abbr} can consistently achieve outstanding performances across heterogeneous datasets and tests, and promote more accurate predictions on more challenging metric $\text{IoU}@0.7$ and mIoU, 
further indicating the effectiveness of the proposed method for enhancing generalizable visual-textual learning.

\vspace{-1mm}
\subsection{Ablation Studies}
\vspace{-1mm}
In this section, we performed in-depth ablations to evaluate
the efficacy of each component in \method{abbr} on the Charades-STA dataset~\cite{gao2017tall}.

\begin{figure}[t]
    \centering
    \begin{minipage}[t]{0.68\textwidth}
    \hspace{-7mm}
        \centering
        \includegraphics[width=1.\linewidth]{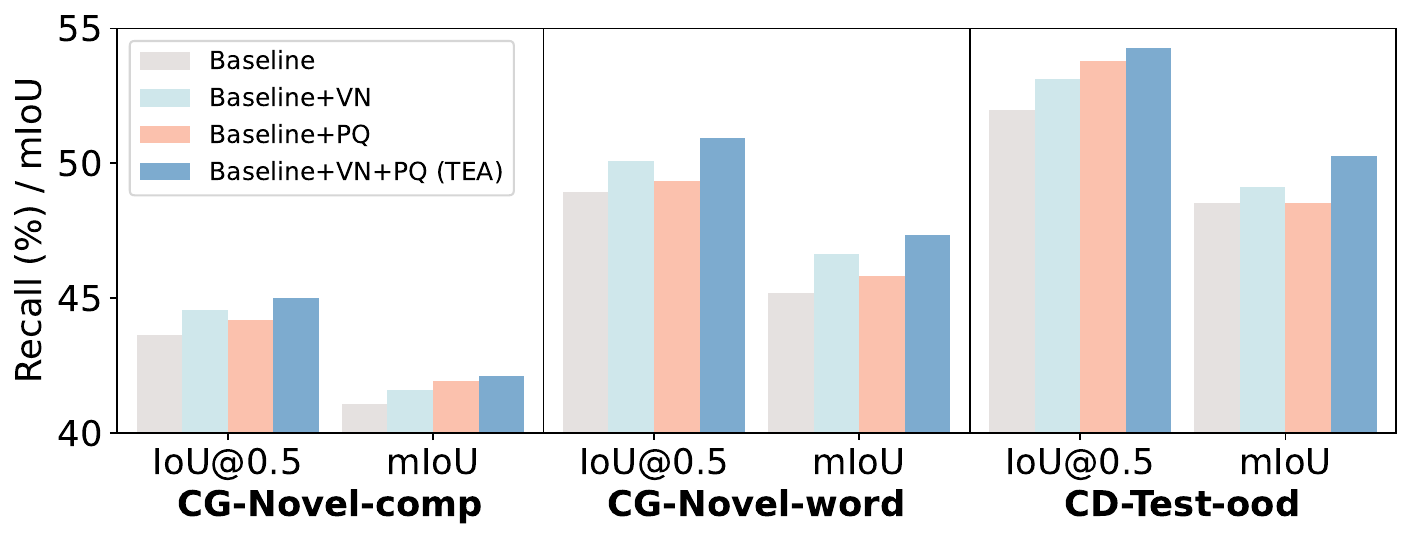}
    \vspace{-3mm}
        \caption{Components analysis}
        \label{fig:abla-component}
    \end{minipage}
\hspace{\fill}
    \begin{minipage}[t]{0.3\textwidth}
        \centering
        \includegraphics[width=0.89\linewidth]{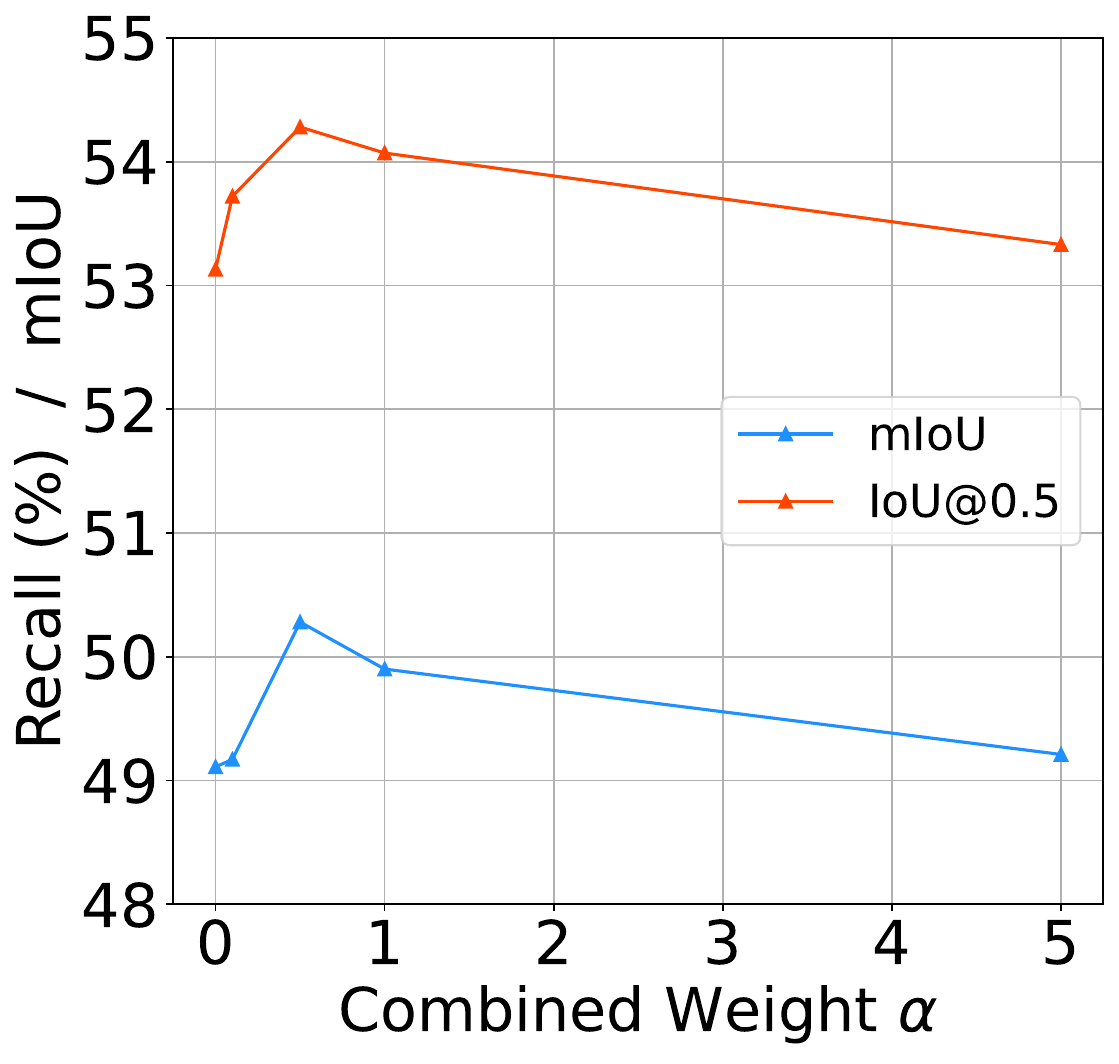}
    \vspace{-3mm}
        \caption{Combined weight}
        \label{fig:abla-weight}
    \end{minipage}
    \hfill
\end{figure}

\vparagraph{Component analysis.}
We examined the effectiveness of each proposed module 
in Fig.~\ref{fig:abla-component}.
The 
video-narrative
knowledge enhancement module is denoted as 
`VN' 
and the paragraph-query parallel interaction module is `PQ'.
We take~\cite{huang2022emb} as the baseline. 
The 
video-narrative
knowledge enhancement  
and paragraph-query parallel interaction modules
proposed in \method{abbr}
have their own benefits to promote more accurate visual-textual understanding.
Moreover, when they are both adopted together 
by text-enhancing the video representations and final predictions jointly, the performance benefited more.

\vparagraph{Different implementations of video-paragraph merging.}
In Eq.~\eqref{eq:merge}, we merge the video and corresponding paragraph knowledge with concatenation and one MLP. Here we discuss other alternative implementation choices:
(a) directly sum the two features (Add):
$s^m_i = \text{Add}(s_i,c_i); V^m = \{s^m_i\}_{i=1}^{L^v} $;
(b) apply the cross-attention like in Eq.~\eqref{eq:attention_step} for interactions (Attention):
$V^m \leftarrow \mathcal{A}(V,P)$.
Table~\ref{tab:abla-merge} 
presents the comparative results on CD-Test-odd, illustrating how employing a weighted combination through concatenation and MLP enables the selective merging of knowledge from both video and paragraph sources during training. This approach maintains distinguishability at each period, thereby facilitating semantic temporal moment localization.

\vparagraph{Different choices to generate text descriptions.}
In \method{abbr}, we use LLaVA \cite{liu2023llava} as the MLLM to generate text descriptions with parallel prompt inputs. There are also some valuable works about captioners without prompts.
Here we select one of the representative models BLIP2~\cite{li2023blip2} 
(\texttt{BLIP2-pretrain-opt6.7b}) as the text generator.
Table~\ref{tab:abla-captioner} shows the comparison results on CD-Test-odd, demonstrating
the effectiveness 
and  {robustness} of \method{abbr} to utilize various description
generators to address the modality imbalance and promote more accurate retrieval.
Given no prompt input in BLIP2, there is no option to adjust the granularity of the output to always maintain distinguishability at each timestamp, and the performance is not as good as promptable text generators.

\begin{figure*}[ht]
\centering
\includegraphics[width=1.01
   \textwidth]{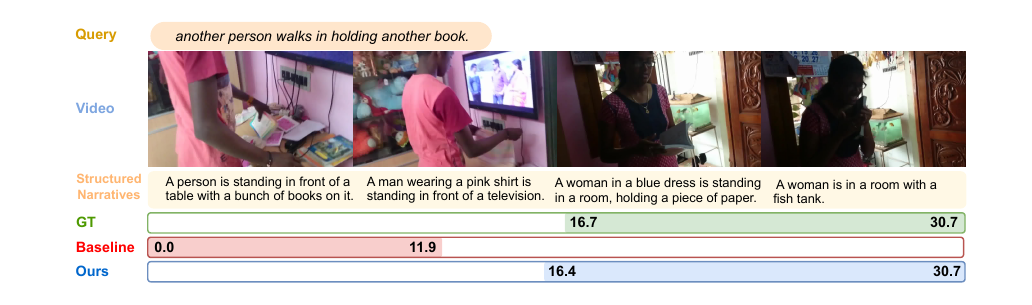}
\vspace{-2mm}
\caption{Qualitative example on Charades. The structured narratives provide guidance (`A man wearing a pink shirt' \vs `A woman in a blue dress') to help the model understand who is the `another person' and get more accurate predictions.}
\label{fig:visualization}
\vspace{-1mm}
\end{figure*}

\vparagraph{Visualization.}
In Fig.~\ref{fig:visualization}, we show an example of prediction from Charades.
For the query \textit{another person walks in holding another book}, 
the model needs to understand who is the \textit{another person} to get the correct prediction.
Baseline fails to understand the semantics of the query and gives an incorrect answer by retrieving one similar action about the first person. When we supplement semantics by the generated structured paragraph, which provides the complementary localization hind (`A man wearing a pink shirt' \vs `A woman in a blue dress'), \method{abbr} can predict a more accurate answer.

\vparagraph{Combination weight in prediction enhancement.}
In Eq.~\eqref{eq:enhanced_predict}, 
we combine the prediction scores 
with the paragraph-query parallel interaction predictions with weight hyper-parameter $\alpha$ to add complemental context information from the text space.
Here, we illustrate the hyper-parameter searching process 
in Fig.~\ref{fig:abla-weight}.
One can see that
with the increment of $\alpha$, the accuracy of the predictions attains its maximum value at 0.5,
and we choose $\alpha$=0.5 as our implementation.

\vspace{-2mm}
\section{Conclusion}
\label{sec:conclusion}
\vspace{-2mm}

In this work, 
we formulated a novel paradigm called \method{full} (\method{abbr}) to solve the 
modality imbalance problem in VMR and enhance generalizable moment-text associations learning.
We leverage an MLLM as a video narrator to
construct structured textual narratives for video content. These
narratives serve as enriched semantic bridges, 
aiding in cross-modal video-query alignment.
Our \method{abbr} model provides state-of-the-art performances on
various different out-of-distribution evaluations from two popular
video moment retrieval benchmark datasets, demonstrating the
effectiveness of the proposed method for enhancing generalizable
visual-text learning. \\
\textbf{Limitations.} From the experiments, it took around 1 second for each frame to get one narrative description from LLaVA. To further facilitate the MLLM's capability more efficiently, the train-free paradigm is a promising research direction in the VMR task.
Moreover, even though we maintained the narratives' robustness through 
neighbor merging
in Eq~\eqref{eq:para_align} in the Temporally Structured Text Paragraph Construction module and also demonstrated the robustness of our design across different MLLMs in Table~\ref{tab:abla-captioner}, designing a more intuitive standard to measure and improve the quality of MLLMs (such as regarding hallucination~\cite{bai2024hallucination}) is a worthwhile direction to be explored in the future.

\bibliographystyle{splncs04}
\bibliography{vmr}

\end{document}